\documentclass[10pt,twocolumn,letterpaper]{article}

\usepackage{cvpr}              

\usepackage{mathtools}
\usepackage{multirow}
\usepackage{arydshln}
\usepackage[table]{xcolor}
\usepackage{amssymb}
\usepackage{graphicx}

\newcommand{\br}[1]{\textcolor{black}{\textbf{#1}}} 
\newcommand{\upr}[1]{%
  \textcolor{red}{$\uparrow$} %
}
\newcommand{\downr}[1]{%
  \textcolor{red}{$\downarrow$} %
}
\newcommand{\mline}[1]{%
  \textcolor{red}{$-$} %
}
\definecolor{cvprblue}{rgb}{0.21,0.49,0.74}
\usepackage[pagebackref,breaklinks,colorlinks,allcolors=cvprblue]{hyperref}

\def\confName{CVPR}
\def\confYear{2026}

\title{Beyond Strict Pairing: Arbitrarily Paired Training for\\ High-Performance Infrared and Visible Image Fusion}

\author{
    Yanglin Deng$^{1}$, 
    Tianyang Xu$^{1}$\thanks{Corresponding author},
    Chunyang Cheng$^{1}$, 
    Hui Li$^{1}$,
    Xiao-jun Wu$^{1}$,
    Josef Kittler$^{2}$ \\
    \noalign{\vspace{3pt}}
    \small $^{1}$ School of Artificial Intelligence and Computer Science, Jiangnan University\\
    \small $^{2}$ Centre for Vision, Speech and Signal Processing (CVSSP), University of Surrey \\
    {\tt\small yanglin\_deng@stu.jiangnan.edu.cn, j.kittler@surrey.ac.uk}\\
    {\tt\small\{tianyang.xu,chunyang\_cheng,lihui.cv, wu\_xiaojun\}@jiangnan.edu.cn} \\
}

\begin{document}

\maketitle
\begin{abstract}
Infrared and visible image fusion(IVIF) combines complementary modalities while preserving natural textures and salient thermal signatures.
Existing solutions predominantly rely on extensive sets of rigidly aligned image pairs for training. 
However, acquiring such data is often impractical due to the costly and labour-intensive alignment process. 
Besides, maintaining a rigid pairing setting during training restricts the volume of cross-modal relationships, thereby limiting generalisation performance.
To this end, this work challenges the necessity of Strictly Paired Training Paradigm (SPTP) by systematically investigating UnPaired and Arbitrarily Paired Training Paradigms (UPTP and APTP) for high-performance IVIF. 
We establish a theoretical objective of APTP, reflecting the complementary nature between UPTP and SPTP.
More importantly, we develop a practical framework capable of significantly enriching cross-modal relationships even with severely limited and unaligned training data.  
To validate our propositions, three end-to-end lightweight baselines, alongside a set of innovative loss functions, are designed to cover three classic frameworks (CNN, Transformer, GAN).
Comprehensive experiments demonstrate that the proposed APTP and UPTP are feasible and capable of training models on a severely limited and content-inconsistent infrared and visible dataset, achieving performance comparable to that of a dataset 100$\times$ larger in SPTP. 
This finding fundamentally alleviates the cost and difficulty of data collection while enhancing model robustness from the data perspective, delivering a feasible solution for IVIF studies. 
The code is available at \href{https://github.com/yanglinDeng/IVIF_unpair}{\textcolor{blue}{https://github.com/yanglinDeng/IVIF\_unpair}}.

\vspace{-5mm}
\end{abstract}    
\newcommand{\rr}[1]{\textcolor{red}{{}#1}}

\section{Introduction}
\label{sec:intro}
\begin{figure}[t]
  \centering
  \includegraphics[width=\linewidth]{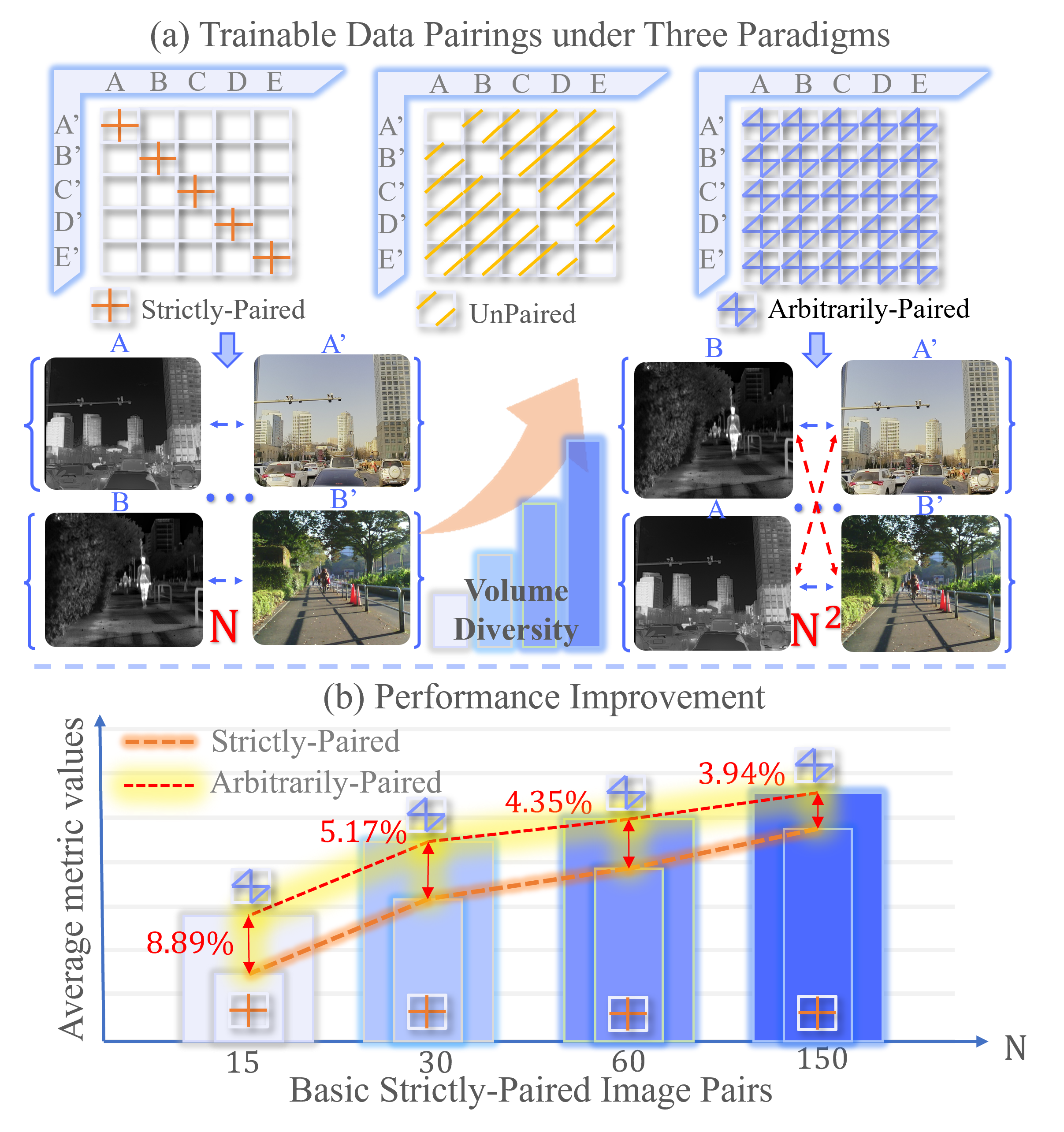}
  \vspace{-6mm}
  \caption{
Motivation of proposed two training paradigms. (a) Compared with the strictly-paired paradigm, the two proposed paradigms lift the requirement for spatiotemporally aligned pairs, expanding the trainable pairs to $N\times(N-1)$ and $N^2$, respectively, reducing data-collection costs and improving generalisation.
(b) Performance improvement pattern: greater enhancement from $N^2$ for Smaller $N$.
  }
  \label{fig:motivation}
  \vspace{-5mm}
\end{figure}

Infrared and visible image fusion (IVIF) aims to generate a comprehensive fused image $O$ that contains both consistent and complementary information from infrared source $I_{ir}$ and visible source $I_{vis}$ \cite{review_1},
\begin{equation}
  O = \mathcal{F}(I_{ir},I_{vis}),
  \label{eq:fusion task}
\end{equation}
where $\mathcal{F}$ represents the fusion model.
Specifically, visible modality excels at capturing colourful visual information aligned with human perception. 
However, it is sensitive to abnormal lighting conditions, occlusion, and poor weather \cite{vis_feature}.
In contrast, infrared modality is robust to illumination variations and environmental challenges ({\em e.g.,} darkness, smoke, haze).
Nevertheless, it lacks colour and fine-grained textural details inherent in the visible modality \cite{ir_fature}.  
Thanks to the complementarity of the two, IVIF has been studied in recent decades with practical applications demands, such as semantic segmentation \cite{seg_1, seg_2, seg_3}, object detection \cite{obj_detect_1, obj_detect_2, obj_detect_3}, and crowd counting~\cite{crowd_count1, crowd_count2}.

In general, an IVIF model is trained to extract the complementary content and consistent scene information across the modalities \cite{densefuse, u2fusion, lrrnet, tgfuse, mufusion, cddfuse, xie2024, song2025, fusionbooster}. 
In principle, fusion performance is determined by four factors, \textit{i.e.}, paired training data, model architecture, supervision loss, and training strategies \cite{review_2}. 
Despite the encouraging performance, existing solutions require a large quantity of strictly aligned infrared and visible image pairs, which poses significant challenges for data collection~\cite{review_1}.
For instance, when collecting strictly paired data in diverse weather, managing spatial and temporal synchronisation and device settings is not trivial.
Besides, considering sensor calibration errors, target motion, and thermal deformations, image registration requires additional effort ~\cite{joint_regisfusion_1,step_regisfusion_1, step_regisfusion_2, step_regisfusion_3}.
Given the inherent differences among modalities, achieving perfect registration is challenging and time-consuming.

Notwithstanding the issues raised by collecting paired image data, the fusion performance is also fragile due to discrepancies in data distributions.
In particular, within a certain dataset, the observed relationships, which are somewhat limited in paired image datasets due to the dataset size, impose an upper bound on the fusion capability that a model can absorb. 
This is a consequence of the well-known result in machine learning that increasing the amount of training data always helps to improve the generalisation performance. 
In conclusion, the conventional strictly-paired training paradigm (SPTP) suffers from critical issues in both cost-effectiveness and model generalisation. 

On the contrary, given the availability of unimodal datasets, it is very easy to create arbitrarily paired infrared and visible pairs.
This convenience motivates us to explore a novel arbitrarily-paired training paradigm (APTP).
In fact, the idea of APTP is quite common in the field of image-to-image generation \cite{cyclegan, unpaired_1, unpaired_2, unpaired_3, unpaired_4, theorm_unpaired}. 
In this particular task, the indispensable conducive factor is the existence of the desired image output. 
Unfortunately, the absence of the ground truth fusion output precludes the application of methods like \cite{cyclegan} to the IVIF task.

To remedy this situation, we pioneer the APTP of IVIF from a brand-new perspective. 
The pixel-level self-supervised nature of image fusion tasks allows us to relax the learning objective of traditional SPTP, 
which is premised on strictly consistent content across modalities. 
Specifically, we explore supervision signals to facilitate arbitrary pixel-level combinations of cross-modal source images, thereby avoiding reliance on strictly consistent content and enabling arbitrary training.
This novel working mechanism motivates the expanded independent relationship between the multi-modal inputs.  
According to its independence, we expand the maximum log-likelihood estimation optimisation objective of fusion system parameters trained from SPTP to APTP. 
Both the independent condition and the improved optimisation formulation of APTP deliver a key message: APTP is the union of the two complementary sets, UPTP (Unpaired Training Paradigm) and SPTP, which inherently poses a broader range of applications. 
Among them, UPTP is the most challenging subtask. 

The unrestricted pairing among source images significantly enhances the trainable cross-modal diversity within a volume-limited base dataset, leading to a more robust optimisation formulation.
This not only fundamentally alleviates the difficulties of data collection and reduces costs, but also enriches the diversity of training data, thereby enhancing model robustness.
To systematically investigate the merits of the proposed APTP and UPTP, we build three concise fusion baselines, including CNN, Transformer, and GAN. 
In particular, CNN aims at extracting context features by focusing on local patterns.
In contrast, the Transformer architecture is expert in long-distance modelling, while ignoring the local details to some extent~\cite{transformerconstrain}.
The generative adversarial network (GAN) baseline follows the generation-and-verification mode to perform network training.
Our experimental comparison of the traditional SPTP, novel UPTP, and APTP, provides convincing support for the proposed innovation. 
Our contributions are as follows:
\begin{itemize}
\item  Two new training paradigms for the IVIF task, which are not constrained to rely exclusively on strictly paired inputs, are presented.
\item A detailed analysis of the data relationships and training requirements across the three paradigms.  
\item  A thorough investigation of the merits of the new training paradigms, which reveals their ability to reduce data-collection cost, to expand the effective training set, and to enhance model robustness. 
\item The experimental results providing supporting evidence for the validity of the proposed APTP and UPTP, and the superiority of their performance compared to SOTA, with high efficiency and robust generalisation. 

\end{itemize}
\section{Related Work}
\label{sec:related}
\subsection{Strictly Paired IVIF}
Most IVIF approaches are trained using strictly paired datasets. 
Depending on the deep learning framework employed, these methods can broadly be categorised into three types: CNN framework \cite{densefuse, rfn-nest, mufusion, mmdrfuse, lrrnet}, CNN-Transformer framework \cite{swinfusion, cddfuse, emma, AFTfusion}, and generative framework \cite{fusiongan, tardal, tgfuse, ddfm}.
To facilitate end-to-end training, \cite{densefuse}, \cite{rfn-nest} introduce dense blocks and residual blocks to explore multi-scale features and avoid the design of hand-crafted fusion rules. 
\cite{mmdrfuse} continues this end-to-end manner of training with a mini-model involving only two CNN blocks, demonstrating that complex network architectures are not necessary for pursuing high performance.
To emphasise the global long-range context, \cite{swinfusion} conducts fusion based on a cross-attention mechanism.
\cite{cddfuse} and \cite{emma} adopt an improved transformer block\cite{restormer} to achieve a trade-off between efficiency and performance.
To discriminate the modality-specific features, \cite{tardal, tgfuse} introduce two discriminators that help to enhance the modality-related clues.
However, GAN-based approaches confront issues such as mode collapse and training instability.
Although \cite{ddfm} applies a likelihood rectification that achieves stable training without a mode collapse, it is computationally heavy.  

In this work, to cover a wide scope in terms of IVIF methodologies, we construct three lightweight baseline models to verify the feasibility, degree of unpairedness, and the merits of APTP and UPTP. 
\subsection{Unpaired Training Mode}
Unpaired training is first introduced in the image generation field to address the prohibitive expense of collecting large-scale paired datasets \cite{pix2pix}. 
\cite{cyclegan} breaks this limitation by introducing adversarial and cycle-consistent supervisions, achieving style transfer with consistent content.  
This innovative paradigm enables training with unpaired data, which greatly broadens task applications. 
Following \cite{cyclegan}, \cite{unpaire_restoration1}, \cite{unpaire_restoration2} frame restoration as an unpaired generation task from degraded to clean images. 
Such challenging problems also exist in the field of IVIF.
Unlike image generation, IVIF demands content consistency between two complementary images rather than within a single one, and lacks a well-defined fused image domain, rendering unpaired training like \cite{cyclegan} infeasible. 
Yet it is precisely the absence of a desired fused image, coupled with the pixel-level nature of the IVIF task, that makes unpaired training still feasible.

Due to the absence of ground-truth, fusion networks are not required to learn from an ideal fused image. 
Instead, supervision is usually drawn from the source images.
Therefore, we argue that the model can still extract the corresponding supervisory signals, even if the input is unpaired. 
Starting from these points, we pioneer in exploring the possibility of training with unpaired source images. 

\section{Approach}
In this section, UPTP and APTP are systematically analysed through theoretical definitions, working mechanisms, and essential properties, respectively. 
Given the lack of prior research, we construct three baselines for comparative verification, which are based on three popular architectures, \textit{i.e.}, CNN, Transformer, and GAN.
The detailed network structures are displayed in the supplementary material.
\subsection{Loss Functions}
Leveraging the pixel-level complementarity of source modalities, we devise loss functions that jointly enforce pixel fidelity, gradient texture, and structural perception. 
Beyond loss items, it is crucial to guide the model in learning to choose suitable source clues and the combination ratio across modalities. 
To enable the model to handle diverse inputs, we design a universal linear weighting function: 
\begin{equation}
    \scalebox{0.8}{$W(a;a,b) = \frac{a}{a+b},
    \label{eq:weighting}$}
\end{equation}
where $a, b$ are the separate inputs to be measured. 
Eqn.~\eqref{eq:weighting} allows the model to automatically decide, at every pixel and in every information scope (intensity, gradient, structure), which source image deserves more attention.  
It preserves the relative magnitudes of source pixels for distortion-free results, thereby stabilising training by preventing gradient explosion and vanishing. 
To leverage this, we design a set of adaptively weighted loss functions. 
\vspace{-1mm}
\begin{gather}
    \left\{
    \scalebox{0.8}{
        $\begin{aligned}
            P_{\mathrm{ir}}   &= W(I_{\mathrm{ir}}; I_{\mathrm{ir}}, I_{\mathrm{vis}}), \\
            P_{\mathrm{vis}}  &= W(I_{\mathrm{vis}}; I_{\mathrm{ir}}, I_{\mathrm{vis}}),
        \end{aligned}$
    }
    \right.
    \left\{
     \scalebox{0.8}{$\begin{aligned}
            G_{\mathrm{ir}}   &= W(\nabla I_{\mathrm{ir}}; \nabla I_{\mathrm{ir}}, \nabla I_{\mathrm{vis}}), \\
            G_{\mathrm{vis}}  &= W(\nabla I_{\mathrm{vis}}; \nabla I_{\mathrm{ir}}, \nabla I_{\mathrm{vis}}),
        \end{aligned}$}
    \right.
    \\
    \left\{
        \scalebox{0.8}{
        $\begin{aligned}
            S_{\mathrm{ir}}   &= W\bigl(\mathit{SSIM}(I_{\mathrm{ir}}, O); \mathit{SSIM}(I_{\mathrm{ir}}, O), \mathit{SSIM}(I_{\mathrm{vis}}, O)\bigr), \\
            S_{\mathrm{vis}}  &= W\bigl(\mathit{SSIM}(I_{\mathrm{vis}}, O); \mathit{SSIM}(I_{\mathrm{ir}}, O), \mathit{SSIM}(I_{\mathrm{vis}}, O)\bigr).
        \end{aligned}
        $}
    \right.
    \\
    \scalebox{0.8}{$
      \begin{aligned}
        &\mathcal{L}_{\mathrm{int}} = \frac{1}{HW} \bigl\| O - (P_{\mathrm{ir}} * I_{\mathrm{ir}} + P_{\mathrm{vis}} * I_{\mathrm{vis}}) \bigr\|_1 \\
        &\mathcal{L}_{\mathrm{grad}} = \frac{1}{HW} \bigl\| \nabla O - (G_{\mathrm{ir}} * \nabla I_{\mathrm{ir}} + G_{\mathrm{vis}} * \nabla I_{\mathrm{vis}}) \bigr\|_1 \\
        &\mathcal{L}_{\mathrm{ssim}} = \bigl\| 1 - \bigl(S_{\mathrm{ir}} * \mathit{SSIM}(I_{\mathrm{ir}}, O) + S_{\mathrm{vis}} * \mathit{SSIM}(I_{\mathrm{vis}}, O)\bigr) \bigr\|_1
        \label{eqn:single_loss}
    \end{aligned}$}
\end{gather}

They collectively form the total loss of the CNN ($\mathcal{L}_{C}$) and Transformer ($\mathcal{L}_{T}$) baselines:
\begin{equation}
\scalebox{0.8}{
$\begin{split}
   \mathcal{L}_{C} =   \mathcal{L}_{T} =  \mathcal{L}_{int}+ \alpha \mathcal{L}_{grad} + \beta \mathcal{L}_{ssim}.
\end{split}$
}
  \label{eq:cnn loss}
\end{equation}
For the GAN baseline, a generator is used to generate the final fused image, which is trained by the adversarial loss ($\mathcal{L}_{adv}$) and the total loss ($\mathcal{L}_{G}$): 
\begin{equation}
   \scalebox{0.8}{$\mathcal{L}_{G} =  \mathcal{L}_{T} =  \mathcal{L}_{int}+ \alpha \mathcal{L}_{grad} + \beta \mathcal{L}_{ssim}- \lambda\mathcal{L}_{adv},$}
\end{equation}
where $\mathcal{L}_{adv}$ is detailed in the supplementary material.

\subsection{Overview of strictly-paired IVIF}
Given the strictly aligned source pairs, infrared image set $X^{ir} \subseteq R^{H \times W \times C}$ and visible image set $X^{vis}\subseteq R^{H \times W \times C}$, where $H$, $W$, $C$ represent the height, width, and the number of channels, the fusion model ($\mathcal{F}$) is expected to deliver an informative fusion output set $O\subseteq R^{H \times W \times C}$. 

By introducing the statistically independent probability of each loss function in Eqn.~\eqref{eqn:single_loss}, we can obtain their joint probability, which is proportional to the optimisation objective of the IVIF task (take the CNN baseline as an example):
\vspace{-2mm}
\begin{equation}
\scalebox{0.85}{
$\begin{split}
     &\underset{\theta}{\arg \max}(\log p_{int}(O|\cdot) +\log p_{grad}(O|\cdot)+\log p_{ssim}(O|\cdot))
     \\
      &\propto \underset{\theta}{\arg \max}\log p(O|\cdot) (\cdot = X^{ir},X^{vis};\theta)  
\end{split}$
}
  \label{eq:loss_pro_ob}
\end{equation}
Therefore, the process of maximising the joint probability is equivalent to minimising the total loss value: 
\vspace{-2mm}
\begin{equation}
\scalebox{0.8}{$\underset{\theta}{arg\ max}\log p(O|\cdot) \Leftrightarrow\underset{\theta}{arg\ min} (\alpha  \mathcal{L}_{int}+\beta \mathcal{L}_{grad} +\mathcal{L}_{ssim}).
  \label{eq:loss_eq_ob}$}
\end{equation}
By introducing the Bayes' theorem, the optimisation objective in Eqn.~\eqref{eq:loss_eq_ob} can be transformed into the combination of a likelihood item and a prior item: 
\begin{equation}
\scalebox{0.8}{$\underset{\theta}{arg\ max}\log p(X^{ir},X^{vis}|O;\theta)+\log p(O;\theta),$}
\end{equation}
where the likelihood measure the probability that the fused images retain from the source, and the prior depicts the image naturalness.
Its element-wise form is as follows: 
\begin{equation}
\scalebox{0.8}{$\underset{\theta  }{arg\ max}\sum_{i=j=1}^{N}(\log p(x_{i}^{ir},x_{j}^{vis}|o_{ij};\theta  ) +\log p(o_{ij};\theta )),
  \label{eq:pair_ob}$}
\end{equation}
where $x_{i}^{ir}\in X^{ir}$, $x_{i}^{vis}\in X^{vis} $, $o_{ij}\in O$.

\subsection{Formulation of Arbitrarily-Paired IVIF}

\label{sec:assumption}
For the conventional SPTP, the model learns to fuse not only the modality characteristics, but also the specifically consistent scene and details. 
In this case, a pair of source inputs should be synchronously captured with the same device setting. 
If there is any misalignment, spatially or temporally, registration is required before or during fusion.


Numerous studies demonstrate that training with strictly paired data is effective, but it is highly costly in labour and time.  
To reduce data collection costs and improve data-use efficiency, we propose
\textbf{ training the fusion model directly on a set of arbitrarily-paired infrared and visible images. }
It can be achieved by directing the fusion model to focus on cross-modal characteristics rather than on rigid content. 
This implies that what the model requires is merely cross-modal data, not necessarily being paired and aligned.  

%

\begin{figure}[h]
  \centering
   \includegraphics[width=1\linewidth]{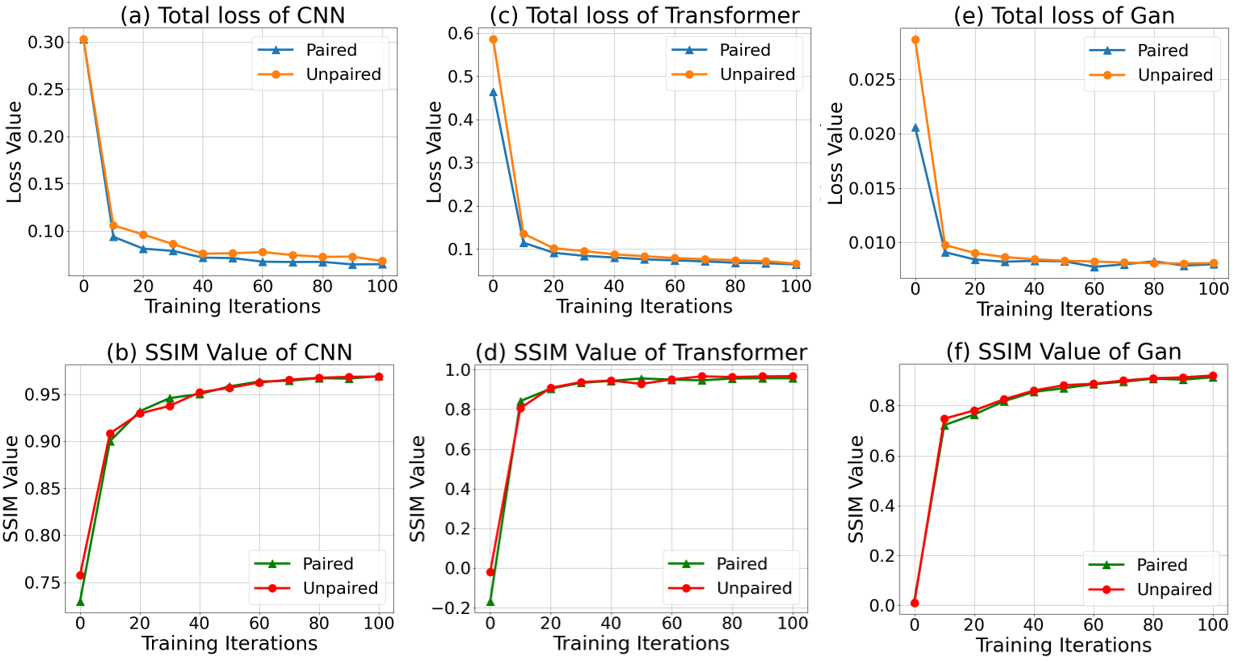}

   \caption{Training convergence and performance comparison between SPTP and UPTP on three baselines. }
   \label{fig:loss}
\vspace{-4mm}
\end{figure}
\begin{figure}[h]
  \centering
   \includegraphics[width=0.95\linewidth]{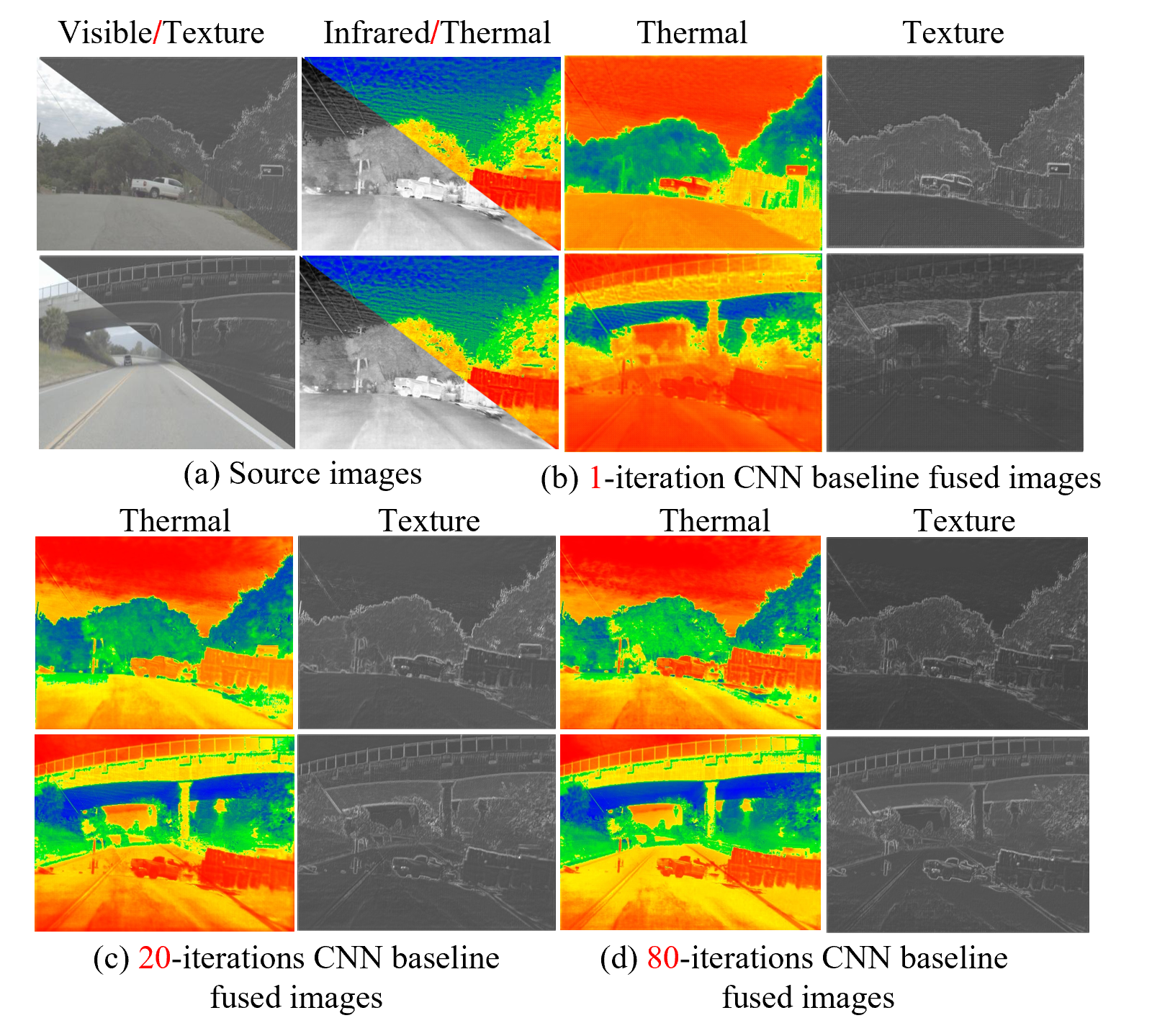}
\vspace{-4mm}
   \caption{Evolution of fusion results during CNN baseline training (SPTP \textit{v.s.} UPTP)  }
   \label{fig:visual_change}
\vspace{-5mm}
\end{figure}
\begin{figure}[h]
  \centering
   \includegraphics[width=0.9\linewidth]{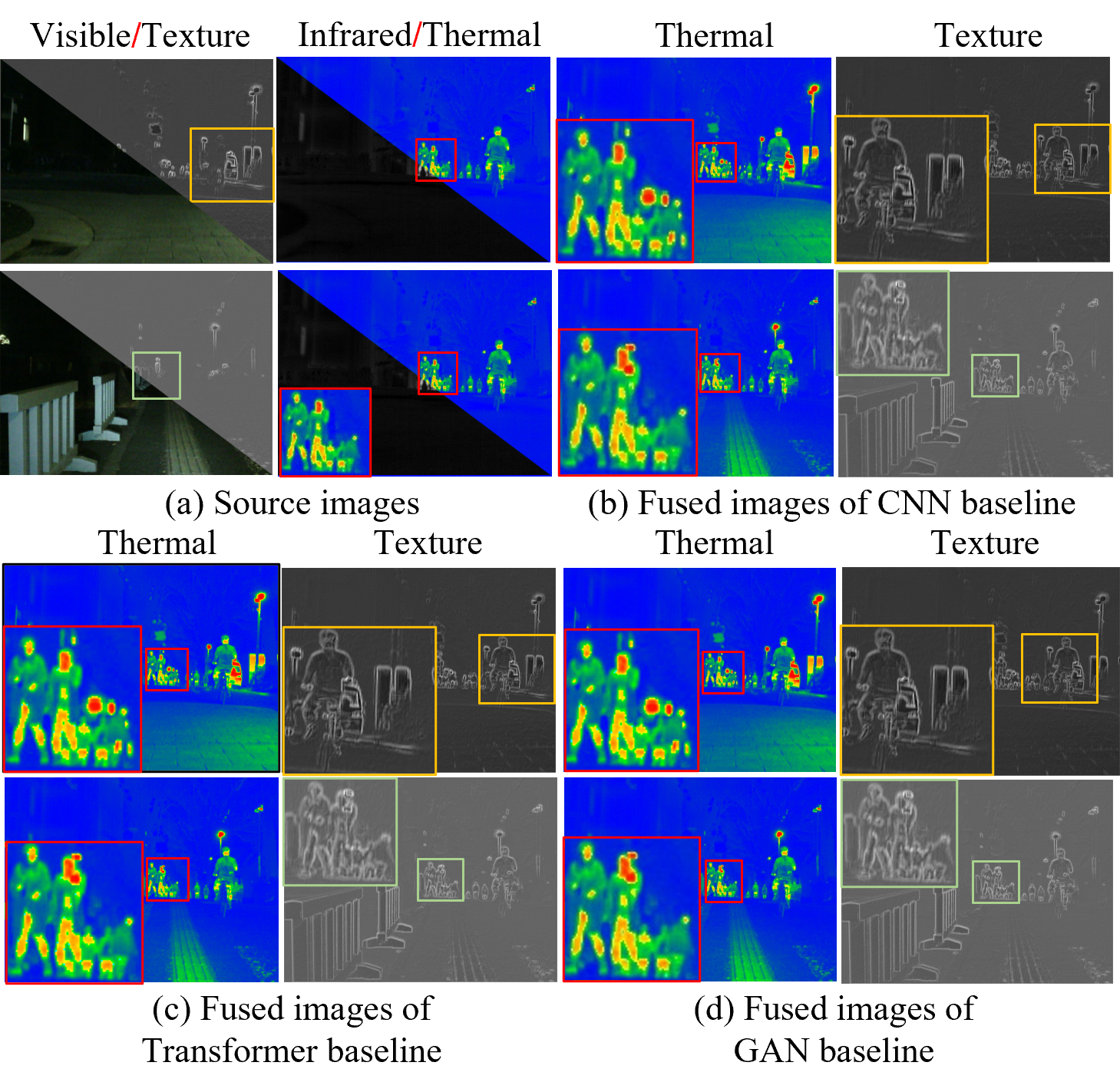}
\vspace{-4mm}
   \caption{Fused images from baselines trained on SPTP and APTP }
   \label{fig:visual_example}
\vspace{-4mm}
\end{figure}
Drawing on this, $x^{ir}_i$ and $x^{vis}_i$ are independent of each other because they can be collected at different locations, at different times, and under different settings.
Mathematically, the pairing relationship in a dataset is random.
Therefore, the joint probability distribution can be decoupled into the product of two individual probability distributions:
\begin{equation}
\scalebox{0.8}{$p(x_{i}^{ir},x_{j}^{vis}) = p(x^{ir}_i)\cdot p(x^{vis}_j).
  \label{eq:independent_form}$}
\end{equation}
Eqn.~\eqref{eq:pair_ob} can be formulated accordingly:
\begin{equation}
\scalebox{0.9}{$\begin{split}
&\underset{\theta \prime }{arg\ max}\sum_{\forall i,j}^{N}(\log p(x_{i}^{ir}|o_{ij}^{\prime};\theta \prime ) +\log p(x_{j}^{vis}|o_{ij}^{\prime};\theta \prime ))
\\
& +\log p(o_{ij}^{\prime};\theta \prime)\propto \underset{\theta \prime}{arg\ max}\sum_{\forall i,j}^{N}\log p(o_{ij}^{\prime}|x_{i}^{ir},x_{j}^{vis};\theta \prime) ,
\label{eq:unpair_ob}
\end{split}$}
\end{equation}
Given Eqn.~\eqref{eq:unpair_ob}, it is worth noting two theoretical observations, \textit{i.e.}, 1. The optimal model parameter set exists when the source inputs $\{x_i^{ir}\}_{i=1}^N$ and $\{x_j^{vis}\}_{j=1}^N$  are independent. 2. The traditional SPTP ($i=j$) and novel UPTP ($i \neq j$) are complementary subsets of the proposed APTP ($\forall i,j$). 
We arrange the data relationship as follows, 
\begin{equation}
\scalebox{0.8}{
$\mathcal{D}_{\text{APTP}}=
\mathcal{D}_{\text{SPTP}}\cup\mathcal{D}_{\text{UPTP}}
\text{with}
\begin{cases}
\mathcal{D}_{\text{SPTP}}=\{(x_i,x_j)\mid i=j\},\\[4pt]
\mathcal{D}_{\text{UPTP}}=\{(x_i,x_j)\mid i\neq j\}.
\end{cases}$
}
  \label{eq:unpair_data}
\end{equation}
It is clear that the proposal of arbitrarily paired training does not aim to overturn all previous image fusion work.
Indeed, it introduces an extended fusion paradigm built upon the SPTP.
Suppose the fusion model and training strategy are not tailored to any specific subset paradigm. 
In that case, they remain applicable to all paradigms, which means that any procedure that suits APTP automatically suits SPTP and UPTP, but the converse does not necessarily hold. 
\begin{figure}[t]
  \centering
   \includegraphics[width=1\linewidth]{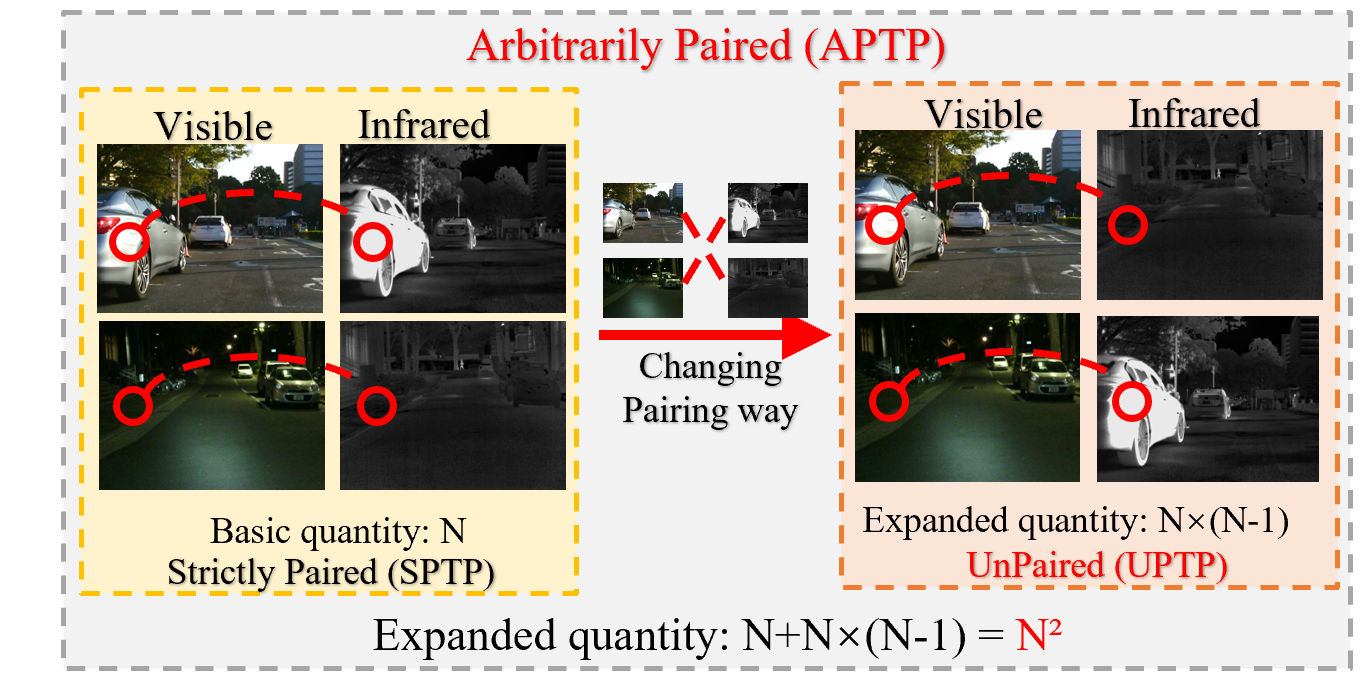}

   \caption{Explanation of relationship diversity.}
   \label{fig:relation}
\vspace{-4mm}
\end{figure}
\begin{figure*}[t]
  \centering
   \includegraphics[width=\linewidth]{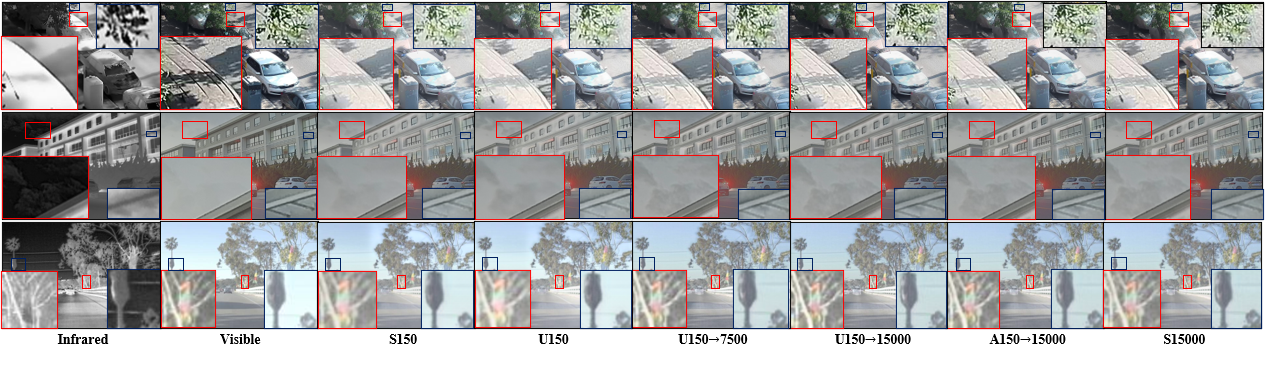}
\vspace{-8mm}
   \caption{Qualitative comparison of three baselines (from top to bottom: CNN, Transformer, and GAN, respectively) under different training data settings on the LLVIP, M3FD, and MSRS test sets.}
   \label{fig:visual1}
\vspace{-4mm}
\end{figure*}
\begin{table*}
  \caption{Qualitative generalisation results on the  LLVIP (250 pairs), MSRS (361 pairs) and M$^3$FD (420 pairs) test sets for three baseline models trained on MSRS with different training settings. Bold fonts indicate higher than their corresponding SPTP baseline groups.}
  \centering
  \label{tab:feasibility}
  \setlength{\tabcolsep}{2pt}
  \footnotesize
  \begin{tabular}{c|c|ccc|ccccc|ccccc|ccccc} 
  \hline 
  \multirow{2}{*}{Datasets} &\multirow{2}{*}{Mode} & \multicolumn{3}{c|}{Quantity} & \multicolumn{5}{c|}{CNN} & \multicolumn{5}{c|}{Transformer} & \multicolumn{5}{c}{GAN} \\ 
  \cline{3-20}
   & & Paired Base & w/o Expanded & Trainable & EN & MI & VIF & Qabf & SSIM & EN & MI & VIF & Qabf & SSIM & EN & MI & VIF & Qabf & SSIM \\ 
  \hline

  \multirow{7}{*}{MSRS} & SPTP & 150 & $\times$ & 150 & 6.54& 2.57 & 0.88 & 0.60 & 0.96 &6.53 & 2.44 & 0.86 & 0.59 & 0.98 & 6.52 & 2.17 & 0.75 & 0.51 & 0.90 \\ 
    & UPTP & 150 & $\times$ & 150 & \br{6.59} & \br{2.62} & \br{0.90} & \br{0.61} & \br{0.97} &  \br{6.55} & \br{2.49} & \br{0.88} & \br{0.60} & 0.98  & \br{6.60} & 2.09 & 0.75 & 0.51 & \br{0.91} \\
    
  & UPTP & 150 & $\checkmark$&7500 & \br{6.56} & \br{2.69} & \br{0.92} & \br{0.65} & \br{0.98} & \br{6.56}& \br{2.69} &\br{0.92} & \br{0.65} & 0.98& \br{6.58}& \br{2.52} & \br{0.90} & \br{0.62} & \br{0.99} \\ 
  
  & UPTP & 150 &$\checkmark$ &15000 & \br{6.57} & \br{2.70} & \br{0.93} & \br{0.65} & \br{0.99} & \br{6.55} & \br{2.70} & \br{0.92} & \br{0.65} & 0.98 & \br{6.58} & \br{2.62} & \br{0.92} & \br{0.63} & \br{0.98} \\ 
  \cline{2-20}
  & SPTP & 15000 & $\times$&15000 & 6.59 & 2.70 & 0.93 & 0.65 & 0.98 & 6.55 & 2.78 & 0.93 & 0.66 & 0.97 & 6.57 & 2.60 & 0.92 & 0.64 & 0.99 \\ 
  & APTP & 150 & $\checkmark$&15000 & 6.58& 2.69 & 0.93 & 0.65 & 0.98 & 6.54& 2.72 & 0.92 & 0.65 & \br{0.98} & 6.57 & 2.60 & 0.92 & 0.63 & \br{1.00} \\ 
  
\hline

  \multirow{7}{*}{LLVIP} & SPTP & 150 & $\times$ & 150 & 7.37 & 2.27 & 0.78 & 0.57 & 0.81 & 7.29 & 2.16 & 0.77 & 0.58 & 0.83 & 7.08 & 1.84 & 0.65 & 0.44&0.72 \\ 
  & UPTP & 150 & $\times$&150 & 7.33 & \br{2.29} & 0.78 & \br{0.59} & \br{0.81} & \br{7.38} & \br{2.18} & \br{0.78} & \br{0.59} & 0.82 & \br{7.33} & \br{1.87} & \br{0.69} & \br{0.46} & \br{0.76} \\
  & UPTP & 150 & $\checkmark$&7500 & 7.31 & \br{2.48} & \br{0.84} & \br{0.63} & \br{0.90} & \br{7.31} & \br{2.41} & \br{0.82} & \br{0.63} & \br{0.89} & \br{7.26} & \br{2.29} &\br{0.76} & \br{0.60}&\br{0.83} \\ 
  & UPTP & 150 & $\checkmark$&15000 & 7.30 & \br{2.54} & \br{0.85}& \br{0.63} & \br{0.91} & \br{7.31} & \br{2.46} & \br{0.85} & \br{0.63} & \br{0.91} & \br{7.29} & \br{2.49} & \br{0.80} & \br{0.60} & \br{0.86} \\ 
  \cline{2-20}
  & SPTP & 15000 & $\times$&15000 & 7.30 & 2.54 & 0.84 & 0.60 & 0.90 & 7.31 & 2.57 & 0.85 & 0.60 & 0.90 & 7.29 & 2.44 & 0.79 & 0.59 & 0.85 \\ 
  & APTP & 150 & $\checkmark$&15000 & 7.30 & 2.54 & 0.84 & \br{0.62}&0.90 & \br{7.32} & 2.50 & 0.85 & \br{0.63} & 0.90 & 7.28 & \br{2.46} & 0.79 & \br{0.60} & 0.84 \\ 
  
  \hline

    \multirow{7}{*}{M$^3$FD} 
 & SPTP & 150 & $\times$&150 & 6.95 & 2.18 & 0.71 & 0.56 & 0.91 & 6.90 & 1.97 & 0.67 & 0.57 & 0.89 & 6.79 & 1.76 & 0.56 & 0.48 & 0.80 \\ 
  & UPTP & 150 & $\times$&150 & 6.93&\br{2.20}&0.71&\br{0.57}&\br{0.93}& 6.90 & \br{2.14} & \br{0.68} & 0.56 & \br{0.90} & \br{6.93} & \br{1.77} & \br{0.59} & 0.48 & \br{0.82} \\
  & UPTP & 150 &$\checkmark$& 7500 & 6.86& \br{2.40} & \br{0.73} & \br{0.58} & \br{0.98} & 6.83 & \br{2.34} & \br{0.72} & \br{0.58} & \br{0.96} & \br{6.87} & \br{2.21} & \br{0.70} &\br{0.57}&\br{0.94} \\ 
  & UPTP & 150 &$\checkmark$& 15000 & 6.83 & \br{2.47} &\br{0.74} & \br{0.58} & \br{0.99} & \br{6.80} & \br{2.43} & \br{0.73} & \br{0.58} & 0.97 & \br{6.90} & \br{2.37} & \br{0.72} & \br{0.57} & \br{0.96}\\ 
  \cline{2-20}
  & SPTP & 15000 & $\times$&15000 & 6.89 & 2.49 & 0.75 & 0.57 & 0.98 & 6.82 & 2.50 & 0.75 & 0.56 & 0.97 & 6.92 & 2.36 & 0.71 & 0.56&0.95 \\ 
  & APTP & 150 & $\checkmark$&15000 & 6.86 & 2.43 & 0.73 & 0.57 & 0.98& 6.81& 2.41 & 0.73 & \br{0.58} & 0.96 & 6.91 & \br{2.37} & \br{0.72} & \br{0.57} & \br{0.96} \\ 

  \hline
  \end{tabular}
  \vspace{-3mm}
\end{table*}
\subsection{Arbitrarily-Paired Training Paradigm}
\label{sec:essence}
 
\subsubsection{Working mechanism of APTP}
In the image fusion field, parameters of a fusion model are continuously updated according to the designed supervision.  
In accordance with the conventional SPTP framework, the loss value is derived from the pixel intensities of images acquired under strictly consistent spatiotemporal conditions.
In contrast, the proposed APTP approach eliminates the need for consistent spatio-temporal imaging settings, thereby enabling supervision from arbitrary cross-modal images via pixel-level supervision. 
However, it does not impede the model from acquiring the required capability, provided that neither the loss functions nor the training strategy are explicitly tailored to enforce cross-modal content consistency (SPTP) or inconsistency (UPTP). 
This is because, in this case, what we use to compute the loss function value is just a desired combination of pixel values, and this combination exhibits different data relationships for different forms of loss functions. 
Therefore, \textbf{even if the trainable images are captured arbitrarily, it does not prevent them from providing trainable pixel relationships to supervise the training of the model.}
As the pixels highlighted by the red circle ~\cref{fig:relation}, different pixel combinations from SPTP and UPTP can provide distinct computational bases for each loss function, whether derived from spatio-temporally aligned or unaligned images. 

The working mechanism can be observed through the loss convergence curves and the progressive changes in visualisation results. 
As the data relationship expressed by Eqn.~\eqref{eq:unpair_data} implies, any method or strategy suited to both SPTP and UPTP is necessarily valid for APTP, therefore, we restrict our comparative analysis to SPTP and UPTP.

In terms of the model convergence, \cref{fig:loss} (a), (c), (e) display the total training loss convergence of three baseline models. 
It is clear that with the same setting, the three baselines exhibit similar convergence behaviour in terms of both speed and final value under both SPTP and UPTP training. 
Theoretically, the fusion performance is expected to improve progressively throughout training. 
To quantitatively assess the integration of luminance, contrast, and structural information, we employ the Structural Similarity (SSIM) index as an evaluation metric. 
As illustrated in \cref{fig:loss} (b), (d), and (f), the SSIM values under both training paradigms exhibit a consistent upward trend on the test set before eventually converging. 
Notably, although UPTP employs unpaired images during training, leading to visually chaotic intermediate fusion results, this does not impede the model's ability to learn meaningful fusion representations. 
When evaluated on paired test images, the outputs are both visually coherent and achieve SSIM scores comparable to those under standard training conditions. 
This observation substantiates our claim that effective fusion models can be learned even in the absence of strictly aligned data.

\cref{fig:visual_change} (b)~(d) present the visualisation of thermal and texture information in the fused images during training, further illustrating the progressive improvement achieved under the UPTP paradigm. 
In the initial training phase, both paradigms struggle to effectively integrate key thermal and texture information. 
Models trained with UPTP, in particular, exhibit visually chaotic outputs due to the lack of explicit pairing. 
As training proceeds with effective supervision, the fusion quality gradually improves, yielding clearer and more distinct thermal targets—such as utility poles, trucks, and trash bins—as well as enhanced texture details like road markings and overpass structures. 
Notably, even at early and intermediate stages (\textit{e.g.}, iterations 20 and 80), the model trained with arbitrarily paired data surpasses the strictly-paired model in certain aspects, displaying more recognisable thermal signatures of objects and sharper texture edges in elements such as roads and utility poles.

To showcase the capabilities of the final trained model, \cref{fig:visual_example}(b)~(d) compare normal as well as abnormal fused images produced by SPTP and UPTP, respectively.
The red boxes show that both of the models trained by SPTP and UPTP can successfully retain the thermal radiation information ({\em e.g.,} people, shining light) contained in the infrared modality. 
The yellow and green boxes demonstrate that they can retain both the texture details and infrared clues ({\em e.g.,} walking people, the person riding a bike, and the bus) contained in the two modalities. 

\subsubsection{Essential advantages}

The inherent randomness of APTP introduces the potential for reconfiguring cross-modal pairing relationships. 
Then, the Eqn.~\eqref{eq:unpair_ob} can be reformulated as follows: 
\begin{footnotesize}
\begin{equation}
\underset{\theta \prime\prime}{arg\ max}\sum_{i=1}^{N}\sum_{j=1}^{N}\log p(o_{ij}^{\prime\prime}|x_{i}^{ir},x_{j}^{vis};\theta \prime\prime) . 
\label{eq:data_scale}
\end{equation}
\end{footnotesize}
Eqn.~\eqref{eq:data_scale} theoretically demonstrates that the base paired datasets $\{x_i^{ir}\}_{i=1}^N$ and $\{x_j^{vis}\}_{j=1}^N$ (with $i=j$) can be expanded to its squared form by pairing all possible image pairs  $\{x_i^{ir}\}_{i=1}^N$ and $\{x_j^{vis}\}_{j=1}^N$ (with $\forall i,j$), yielding a ratio of paired to unpaired data of $1:(N-1)$.  
In principle, the proposed method offers three key advantages: (1) It enables effective scaling of base datasets even when initial data volumes are limited; (2) It enriches intra-dataset relationships to a certain degree, thereby improving model robustness;
(3) Data acquisition costs are significantly reduced, particularly for rare data collected under challenging conditions.

\begin{figure}[h]
  \centering
   \includegraphics[width=1\linewidth]{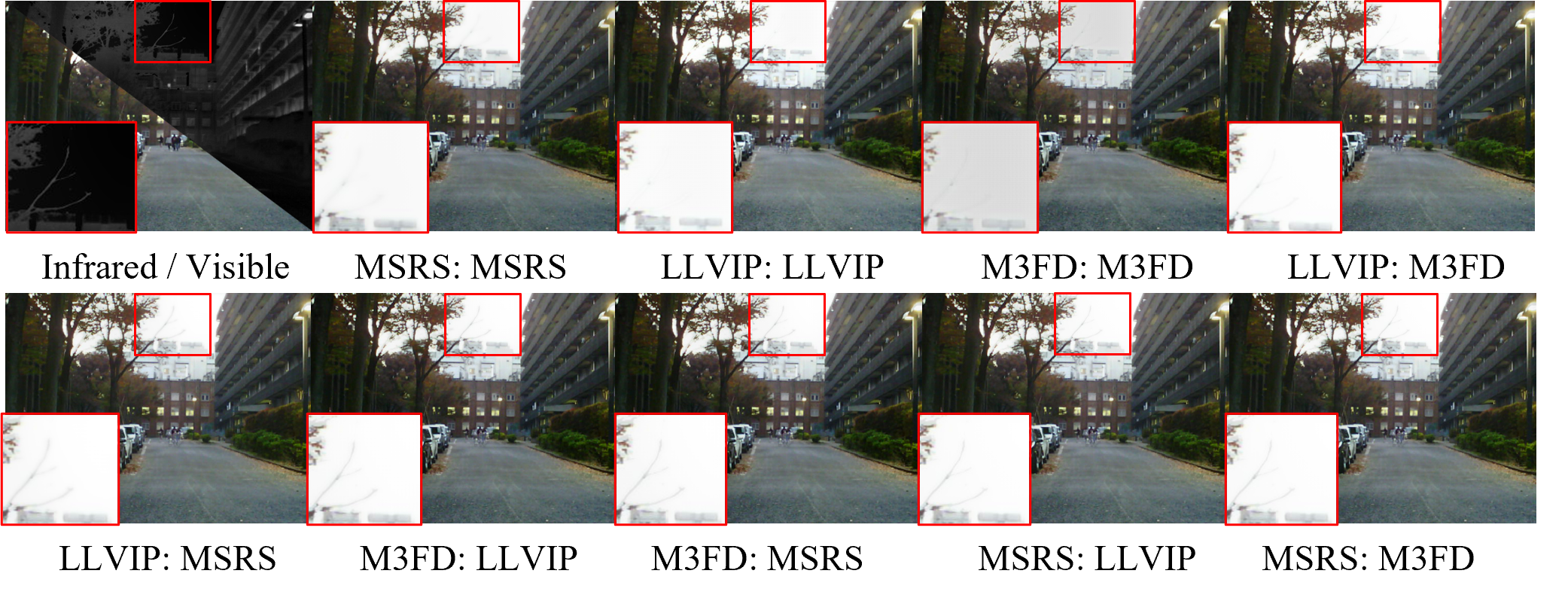}
\vspace{-4mm}
   \caption{Visualisation of fused images: cross-dataset unpaired training \textit{v.s.} single-dataset strictly-paired training models. 
   }
   \label{fig:cross}
\vspace{-4mm}
\end{figure}

As outlined in \cref{sec:intro}, under the traditional strictly-paired training paradigm (SPTP), the representational capacity of a model is constrained by the limited cross-modal relationships present in the dataset. 
In contrast, as shown in \cref{fig:relation}, the arbitrarily-paired training paradigm (APTP) enables the model to capture a quadratically growing set of relationships due to its flexible cross-modal matching mechanism.
Consequently, models trained with APTP exhibit stronger generalisation capabilities. 
While real strictly-paired datasets may contain more images, the learnable relationships remain constrained by their fixed pairing structure.
Through APTP, the model can derive novel relational patterns beyond those available in the original data, thereby enhancing its ability to generalise.

\begin{table}
  \caption{Qualitative results on MSRS of CNN baseline trained on different training sets. Bold indicates that the model performs better than the one trained on its original paired data.}
  \centering
  \label{tab:cross_data}
  \setlength{\tabcolsep}{2pt} 
  \footnotesize
  \begin{tabular}{c|ccc|c|c|ccccc}
    \hline
    \multirow{2}{*}{Mode}&\multicolumn{3}{c|}{Quantity} & \multirow{2}{*}{Infrared} & \multirow{2}{*}{Visible} &  \multirow{2}{*}{EN} &  \multirow{2}{*}{MI} &  \multirow{2}{*}{VIF} &  \multirow{2}{*}{Qabf} &  \multirow{2}{*}{SSIM} \\
    \cline{2-4} 
     &PB & w/o E & T & & &  &  &  &  &  \\
    \hline
    \multirow{3}{*}{SPTP}&150 & $\times$ & 150 & MSRS & MSRS & 6.54 & 2.57 & 0.88 & 0.60 & 0.96 \\
    
    &150 & $\times$ & 150 & M3FD & M3FD & 6.52 & 2.37 & 0.84 & 0.58 & 0.90 \\
  
    &150 & $\times$ & 150 & LLVIP & LLVIP & 6.63 & 2.43 & 0.90 & 0.61 & 0.99 \\
    \hline
    \multirow{6}{*}{UPTP}
    &150 & $\checkmark$ & 7500 & M3FD & MSRS & \br{6.57}& \br{2.66}& \br{0.93}& \br{0.66}& \br{1.01}\\
   
    &150 & $\checkmark$ & 7500 & M3FD & LLVIP & \br{6.59}& \br{2.63}& \br{0.93}& \br{0.65}& \br{1.01}\\

    &150 & $\checkmark$ & 7500 & LLVIP & MSRS & 6.54 & \br{2.74}& \br{0.92}& \br{0.65}& \br{1.00}\\

    &150 & $\checkmark$ & 7500 & LLVIP & M3FD & 6.51 & \br{2.63}& \br{0.91}& \br{0.62}& \br{0.95}\\
    &150 & $\checkmark$ & 7500 & MSRS & LLVIP & 6.54& \br{2.64}& \br{0.91}& \br{0.63}& 
    \br{0.97}\\
 &150 & $\checkmark$ & 7500 & MSRS & M3FD & 6.49 & \br{2.59}& \br{0.89}& \br{0.61}& 0.84 \\
    \hline
  \end{tabular}
  \vspace{-4mm}
\end{table}

\begin{table}[t]
\centering
\captionsetup{skip=2pt}
\caption{Quantitative results on MSRS of comparing the improvement of APTP on different basic strictly-paired images.} 
\label{tab:change}
\setlength{\tabcolsep}{2pt}
 \footnotesize
\begin{tabular}{c|ccc|ccccc|c}

\hline
\multirow{2}{*}{ Mode}&\multicolumn{3}{c|}{Quantity} & \multirow{2}{*}{EN}& \multirow{2}{*}{MI}& \multirow{2}{*}{VIF}  &\multirow{2}{*}{Qabf}&\multirow{2}{*}{SSIM}&\multirow{2}{*}{Avg. Imp.} \\
\cline{2-4}
& PB & w/o E & T& &&&&& \\
\hline

 SPTP& 15&$\times$&15& 6.29&  2.25& 0.74&0.52&0.65&$-$\\
   APTP&15&$\checkmark$&225&6.45& 2.56& 0.80&0.56&0.73&\br{8.89\%}\\
\hline
SPTP& 30&$\times$&30& 6.48&  2.48& 0.80 &0.59&0.93&$-$\\
APTP&30&$\checkmark$&900& 6.55& 2.62& 0.90&0.61&0.96&\br{5.17\%}\\
\hline
SPTP& 60&$\times$&60& 6.54& 2.57& 0.88&0.60&0.96&$-$\\
 APTP&60&$\checkmark$&3600&6.56&2.68&0.93&0.65&0.99&\br{4.35\%}\\
\hline
SPTP& 150&$\times$&150& 6.61& 2.60& 0.91&0.63&0.98&$-$\\
 APTP&150&$\checkmark$&22500&6.58&2.70&0.95&0.68&1.01&\br{3.94\%}\\
\hline
\end{tabular}
\vspace{-5mm}
\end{table}

\begin{table*}
  \caption{Qualitative results of three baseline models and several SOTA methods. Bold shows the best value.}
  \centering
  \label{tab:compare}
  \setlength{\tabcolsep}{2pt} 
  \footnotesize
  \begin{tabular}{c|c|c|ccccc|ccccc|ccccc|ccccc} \hline
       
&\multirow{2}{*}{Venue}&\multirow{2}{*}{MB}& \multicolumn{5}{c}{LLVIP}& \multicolumn{5}{c}{MSRS}& \multicolumn{5}{c}{RoadScene} &\multicolumn{5}{c}{TNO}\\ \cline{4-23}
  & &&EN& MI&VIF&Qabf&SSIM& EN& MI& VIF&Qabf&SSIM& EN& MI& VIF&Qabf&SSIM & EN& MI& VIF&Qabf&SSIM\\ \hline 
  
 DeF \cite{defusion}& 22' ECCV& 30.04&7.25&  2.33& 0.74& 0.43& 0.83& 3.34&  2.15& 0.75& 0.51& 0.93& 6.80& 2.08& 0.50& 0.38&0.86 & 6.94&1.81 &0.53 &0.34 &0.91\\  
 LRR \cite{lrrnet}& 23' TPAMI& 0.19&6.67&  1.65& 0.56& 0.47& 0.66& 6.19&  2.03& 0.54& 0.45& 0.71& 7.12& 2.09& 0.51& 0.38&0.71 & 6.90&1.74 &0.56 &0.37 &0.83\\ 
 Meta \cite{metafusion}& 23' CVPR& 3.10&7.14&  1.16& 0.61& 0.29& 0.60& 6.37&  1.16& 0.71& 0.48& 0.78& 7.17& 1.60& 0.51& 0.36&0.75 &\br{7.09} &1.12 &0.57 &0.35 &0.80\\ 
 DDF \cite{ddfm}& 23' ICCV& 2108&7.17&  1.78& 0.58& 0.23& 0.58& 6.17& 1.89 &0.74 &0.47 &0.91 & 7.42& 2.13& 0.57& 0.42&\br{0.88} &6.83 &1.46 &0.56 &0.36 &0.90\\ 
 Cross \cite{crossfuse}& 24' Inf.& 79.40&6.47&  1.79& 0.69& 0.44& 0.71& 6.49&  2.17& 0.84& 0.56& 0.88& 7.25& 2.13& 0.58& 0.39&067 &6.83 &2.10 &0.74 &0.45 &0.88\\ 
 CTHI \cite{cthie}& 24' IJCV& 53.89&6.83&  1.44& 0.33& 0.22& 0.40& 5.64& 1.45& 0.43& 0.35& 0.47& 7.23& 1.65& 0.36& 0.30&0.46 &6.97 &1.29 &0.41 &0.31 &0.56\\ 
 DCINN \cite{dcinn}& 24' IJCV& 9.77&7.04&  2.00& 0.75& 0.52& 0.77& 6.00& 2.30& 0.82& 0.57& 0.73& 6.90& 1.77& 0.23& 0.23&0.33 &6.67 &1.58 &0.42 &0.33 &0.59\\ 
 Free \cite{freefusion}& 25' TPAMI& 65.19&7.09&  1.77& 0.76& 0.54& 0.68& 5.16&  1.38& 0.51& 0.42& 0.43& 6.70& 1.73& 0.51& 0.37&0.73 & 7.07&1.35 &0.55 &0.35 &0.80\\
 SAGE \cite{sage}& 25' CVPR& 0.55&6.96&  1.81& 0.73& 0.57& 0.81& 6.00& 2.23& 0.71& 0.54& 0.88& 6.76& 2.17& 0.53& 0.39&0.86 &6.91 &1.78 &0.63 &0.46 &0.97\\ 
 GIFNet \cite{gifnet}& 25' CVPR& 3.25& 7.01&1.56& 0.64& 0.47& 0.74& 5.94&1.36& 0.58& 0.42& 0.85& 7.14& 1.86& 0.48& 0.37&0.84 &6.83 &1.34 &0.49 &0.36 &0.88\\ \hline
 CNN&Ours&0.81& \br{7.34}&\br{2.54}& \br{0.88}& \br{0.63}& \br{0.91}& \br{6.58}&\br{2.69}& \br{0.93}& \br{0.65}& \br{1.00}& \br{7.45}&\br{2.42}& \br{0.66}& \br{0.47}&0.87 &6.93 &\br{2.23}&\br{0.77}&\br{0.53}&\br{0.98}\\ 
 Transf& Ours& 0.81&\br{7.35}& \br{2.56}& \br{0.87}& \br{0.64}& \br{0.92}& \br{6.57}& \br{2.64}& \br{0.92}& \br{0.65}& \br{0.99}& \br{7.44}& \br{2.38}& \br{0.65}& \br{0.47}&0.86 &6.94&\br{2.18}&\br{0.76}&\br{0.52}&\br{0.99}\\ 
  GAN&Ours& 0.71& \br{7.35}& \br{2.51}& \br{0.87}& \br{0.65}& \br{0.92}& \br{6.58}& \br{2.56}& \br{0.92}& \br{0.63}& \br{1.01}& \br{7.46}&  \br{2.36}& \br{0.65}& \br{0.46}&\br{0.88} &6.96&\br{2.13}&\br{0.76}&\br{0.52}&\br{0.99}\\ \hline
\end{tabular}
\vspace{-4mm}
\end{table*}

\section{Experiments}

\subsection{Setup}
The evaluation datasets include MSRS \cite{msrs}, M$^3$FD \cite{tardal}, LLVIP \cite{llvip}, RoadScene \cite{road}, and TNO \cite{tno}. 
In terms of qualitative evaluation metrics, we adopt one non-reference metric: entropy (EN), as well as four full-reference metrics: mutual information (MI), visual information fidelity (VIFF),  $\mathrm{Q^{AB/F}}$ and structural similarity index measure (SSIM). 
These metrics can comprehensively assess the image quality both from the fused image itself and from the differences between the fused image and the source images, as described in \cite{metrics}. 
Our evaluations are conducted on a Linux server with an NVIDIA GeForce RTX 3090 GPU.
The hyperparameter settings for all experiments are the same, including $\alpha$=1, $\beta$=0.2, and $\lambda$=0.01.

\subsection{Validation Experiments}
\subsubsection{Feasibility of UPTP and APTP}
With the relaxed optimisation objective, the model learns pixel-to-pixel relationships across modalities, making both strictly-paired training (SPTP) and unpaired training (UPTP) feasible. 
To validate this, we train three baseline fusion models using 150 paired and 150 unpaired images from the MSRS dataset, and evaluate their fusion performance qualitatively and quantitatively on the M$^3$FD, MSRS, and LLVIP benchmarks.
As reported in Table~\ref{tab:feasibility} (rows 1–2 in each group), the fusion results under SPTP and UPTP are nearly identical across five evaluation metrics on all three datasets. 
In the Transformer-based baseline, UPTP even achieves marginally better performance. 
This observation is visually supported in Figure~\ref{fig:visual1}, where results in the fourth column preserve richer modality-specific details in certain elements (\textit{e.g.}, clouds, roadside branches, and road surfaces) while maintaining comparable clarity in others (\textit{e.g.}, the light bulb), relative to the third column.

These outcomes indicate that, through the proposed UPTP strategy, models can leverage randomly paired cross-modality samples to capture data relationships that are comparable to, or even richer than, those learned from an equivalent volume of strictly-paired data.

\subsubsection{Advantages of APTP}
\noindent\textbf{Enriching Data Relationship within Single Datasets}
Leveraging the data scalability benefit formalised in Eqn.~\eqref{eq:data_scale} (\textbf{Advantage 1}), we construct unpaired training sets by expanding the original 150 image pairs from the MSRS dataset to 50$\times$ and 100$\times$ of its initial size. 
By recombining source images, we enrich the diversity of cross-modal relationships available during training, thereby enhancing the generalisation capability of the fusion models.
This is corroborated by the consistent improvement in evaluation metrics across all three datasets and baseline models in Table~\ref{tab:feasibility} (rows 1$\sim$4 in each group). 
It is noted that in some cases, the EN value decreases as the amount of trainable data increases, while other metrics continue to improve. 
This suggests that the model is filtering out noise, producing fusion results with higher information density, better visual quality, and more structurally coherent outputs. 
Further visual evidence is provided in Figure~\ref{fig:visual1}, where regions highlighted in the third to sixth columns show progressively clearer and more detailed outputs.

To examine the robustness enhancement enabled by UPTP and APTP, corresponding to \textbf{Advantage 2} in Eqn.~\eqref{eq:data_scale}, we design a comparative study using three training sets, each containing 15,000 image pairs expanded from the same base set of 150 paired images. 
These sets are constructed under strictly-paired (SPTP), completely unpaired (UPTP), and arbitrarily-paired (APTP) protocols. 
As shown in Figure~\ref{fig:visual1} and Table~\ref{tab:feasibility} (rows 4$\sim$6 in each group), all three baselines achieve nearly identical qualitative performance across paradigms. 
This result underscores a key advantage of the proposed approach, \textit{i.e.}, by strategically re-pairing existing data, we can match the fusion performance of models trained with 100$\times$ more strictly-paired data, significantly improving data utilisation and training flexibility.

\noindent\textbf{Enriching Data Relationship within Cross Datasets}
In terms of the \textbf{Advantage 3} of Eqn.~\eqref{eq:data_scale}, we exploit three datasets to conduct unpaired training across them. 
Despite the large differences in scene content across datasets, they still fall within the applicability of our proposed training paradigms, admitting the fact that a fusion model learns content-independent pixel-to-pixel relationships. 

As shown in the Table~\ref{tab:cross_data}, different pixel combinations across data pairs present distinct learnable information. 
Among the configurations, the first five provide sufficient effective training relationships, showing consistent improvements in MI, VIF, Qabf, and SSIM, along with only minor EN fluctuations, demonstrating enhanced fusion efficiency, structural fidelity, visual quality, and noise suppression. 
In contrast, the last group shows gains in MI, VIF, and Qabf but drops in EN and SSIM, indicating a trade-off that enhances salient targets and edges at the cost of natural appearance and texture detail, making it a suboptimal choice. 
The overall improvement in visual quality and clearer branch details in the Figure~\ref{fig:cross} provide compelling evidence for this conclusion.

\noindent\textbf{Enriching Data Relationships from Datasets of Varying Pairing Quantities}
Based on the constructed baseline under capacity constraints, we design different amounts of basic strictly-paired data, \textit{i.e.}, 15, 30, 60 and 150 image patches.
Using the proposed APTP with varying pairing orders, we expanded the maximum trainable data volume to 225, 900, 3600, and 22500, respectively, to explore the potential gain of APTP. 
Since the built baseline model size is only 0.88MB, excessively increasing the trainable data volume adversely affects model performance.
As shown in the Table~\ref{tab:change}, APTP can consistently enhance model robustness by enriching trainable data relationships. 
Notably, the relative improvement achieved by the proposed APTP is more pronounced when the basic data volume is smaller.
Therefore, in real-world scenarios, we can pursue satisfactory model performance with only a small amount of collected data, significantly enhancing task efficiency.

\subsection{Comparison with State-of-the-Art Methods}
Motivated by the outstanding cross-dataset performance of MSRS and M$^3$FD in Table~\ref{tab:cross_data}, we use 150 paired images from MSRS and M$^3$FD as visible and infrared source images, respectively, and expand the training set to 15,000 pairs to train three baseline models. 
The resulting models are compared with state-of-the-art (SOTA) methods to demonstrate the superiority of the proposed training paradigms and loss functions. 
To ensure fairness, the details of the training data used for all methods are provided in the supplementary material. 
As clearly shown in the Table~\ref{tab:compare}, our methods achieve superior performance on all four datasets. 
Notably, its excellence on three unseen datasets strongly demonstrates a powerful generalisation capability and the effectiveness of the proposed loss functions. 
Despite using only 150 image pairs, our approach of varying pairing strategies creates rich, trainable pixel relationships for all models. 
This not only enhances robustness but also fundamentally reduces the model's dependence on large data volumes, thereby significantly cutting data collection costs. 
Additional visual results and corresponding analyses are provided in the supplementary material. 

In addition to the experimental investigations presented above, we provide further analysis of the proposed training paradigms in the supplementary material.

       
 

\section{Conclusion}
This work challenges the necessity of the Strictly Paired Training Paradigm (SPTP) in infrared and visible image fusion by systematically exploring UnPaired and Arbitrarily Paired Training Paradigms (UPTP and APTP). Theoretically, UPTP and SPTP are shown to be complementary subsets of the more flexible APTP. The proposed solution significantly expands the volume and diversity of cross-modal relationships from a limited dataset. Through three lightweight baselines and novel loss functions, experiments demonstrate that UPTP and APTP can train models on severely limited, content-inconsistent data, achieving performance comparable to models trained on a dataset 100 times larger under SPTP. This finding substantially reduces data collection costs and enhances model robustness from a data perspective. 

\section{Acknowledgement}
This work was supported in part by the National Natural Science Foundation of China (62576152, 62332008, 62336004), the Basic Research Program of Jiangsu (BK20250104), the Fundamental Research Funds for the Central Universities (JUSRP202504007), and Leverhulme Trust Emeritus Fellowship EM-2025-06-09.

{
    \small
    \bibliographystyle{ieeenat_fullname}
    \bibliography{main}

@String(PAMI = {IEEE Trans. Pattern Anal. Mach. Intell.})

@String(IJCV = {Int. J. Comput. Vis.})

@String(CVPR= {IEEE Conf. Comput. Vis. Pattern Recog.})

@String(ICCV= {Int. Conf. Comput. Vis.})

@String(ECCV= {Eur. Conf. Comput. Vis.})

@String(TIP  = {IEEE Trans. Image Process.})

@String(IJCAI = {IJCAI})

@String(AAAI = {AAAI})

@String(PAMI  = {IEEE TPAMI})

@String(IJCV  = {IJCV})

@String(CVPR  = {CVPR})

@String(ICCV  = {ICCV})

@String(ECCV  = {ECCV})

@String(TIP   = {IEEE TIP})

@article{review_1,
  author       = {Xingchen Zhang and
                  Yiannis Demiris},
  title        = {Visible and Infrared Image Fusion Using Deep Learning},
  journal      = {{IEEE} PAMI},
  volume       = {45},
  number       = {8},
  pages        = {10535--10554},
  year         = {2023},
}

@article{review_2,
  author       = {Shahid Karim and
                  Geng Tong and
                  Jinyang Li and
                  Akeel Qadir and
                  Umar Farooq and
                  Yiting Yu},
  title        = {Current advances and future perspectives of image fusion: {A} comprehensive
                  review},
  journal      = {Inf. Fusion},
  volume       = {90},
  pages        = {185--217},
  year         = {2023},
}

@inproceedings{ir_fature,
  author       = {Zixiang Zhao and
                  Jiangshe Zhang and
                  Xiang Gu and
                  Chengli Tan and
                  Shuang Xu and
                  Yulun Zhang and
                  Radu Timofte and
                  Luc Van Gool},
  title        = {Spherical Space Feature Decomposition for Guided Depth Map Super-Resolution},
  booktitle    = {ICCV},
  pages        = {12513--12524},
  publisher    = {{IEEE}},
  year         = {2023},
}

@inproceedings{vis_feature,
  author       = {Hao Zhang and
                  Linfeng Tang and
                  Xinyu Xiang and
                  Xuhui Zuo and
                  Jiayi Ma},
  title        = {Dispel Darkness for Better Fusion: {A} Controllable Visual Enhancer
                  Based on Cross-Modal Conditional Adversarial Learning},
  booktitle    = {CVPR},
  pages        = {26477--26486},
  publisher    = {{IEEE}},
  year         = {2024},
}

@inproceedings{step_regisfusion_1,
  author       = {Prune Truong and
                  Martin Danelljan and
                  Radu Timofte},
  title        = {GLU-Net: Global-Local Universal Network for Dense Flow and Correspondences},
  booktitle    = {CVPR},
  pages        = {6257--6267},
  publisher    = {{IEEE}},
  year         = {2020},
}

@inproceedings{step_regisfusion_2,
  author       = {Moab Arar and
                  Yiftach Ginger and
                  Dov Danon and
                  Amit H. Bermano and
                  Daniel Cohen{-}Or},
  title        = {Unsupervised Multi-Modal Image Registration via Geometry Preserving
                  Image-to-Image Translation},
  booktitle    = {CVPR},
  pages        = {13407--13416},
  publisher    = {{IEEE}},
  year         = {2020},
}

@inproceedings{step_regisfusion_3,
  author       = {Di Wang and
                  Jinyuan Liu and
                  Xin Fan and
                  Risheng Liu},
  title        = {Unsupervised Misaligned Infrared and Visible Image Fusion via Cross-Modality
                  Image Generation and Registration},
  booktitle    = {IJCAI},
  pages        = {3508--3515},
  publisher    = {ijcai.org},
  year         = {2022},
}

@inproceedings{joint_regisfusion_1,
  author       = {Han Xu and
                  Jiayi Ma and
                  Jiteng Yuan and
                  Zhuliang Le and
                  Wei Liu},
  title        = {RFNet: Unsupervised Network for Mutually Reinforcing Multi-modal Image
                  Registration and Fusion},
  booktitle    = {CVPR},
  pages        = {19647--19656},
  publisher    = {{IEEE}},
  year         = {2022},
}

@inproceedings{transformerconstrain,
  author       = {Zhendong Wang and
                  Xiaodong Cun and
                  Jianmin Bao and
                  Wengang Zhou and
                  Jianzhuang Liu and
                  Houqiang Li},
  title        = {Uformer: {A} General U-Shaped Transformer for Image Restoration},
  booktitle    = {CVPR},
  pages        = {17662--17672},
  publisher    = {{IEEE}},
  year         = {2022},
}

@inproceedings{seg_1,
  title={PIDNet: A real-time semantic segmentation network inspired by PID controllers},
  author={Xu, Jiacong and Xiong, Zixiang and Bhattacharyya, Shankar P},
  booktitle={CVPR},
  pages={19529--19539},
  publisher= {{IEEE}},
  year={2023}
}

@inproceedings{seg_2,
  title={Federated incremental semantic segmentation},
  author={Dong, Jiahua and Zhang, Duzhen and Cong, Yang and Cong, Wei and Ding, Henghui and Dai, Dengxin},
  booktitle={CVPR},
  publisher = {{IEEE}},
  pages={3934--3943},
  year={2023}
}

@article{seg_3,
  author       = {Tianfei Zhou and
                  Wenguan Wang},
  title        = {Cross-Image Pixel Contrasting for Semantic Segmentation},
  journal      = {{IEEE} PAMI},
  volume       = {46},
  number       = {8},
  pages        = {5398--5412},
  year         = {2024},
}

@inproceedings{obj_detect_1,
  title={Camouflaged object detection with feature decomposition and edge reconstruction},
  author={He, Chunming and Li, Kai and Zhang, Yachao and Tang, Longxiang and Zhang, Yulun and Guo, Zhenhua and Li, Xiu},
  booktitle={CVPR},
  publisher = {{IEEE}},
  pages={22046--22055},
  year={2023}
}

@inproceedings{obj_detect_2,
  title={Adamixer: A fast-converging query-based object detector},
  author={Gao, Ziteng and Wang, Limin and Han, Bing and Guo, Sheng},
  booktitle={CVPR},
  publisher = {{IEEE}},
  pages={5364--5373},
  year={2022}
}

@article{obj_detect_3,
  author       = {Gong Cheng and
                  Xiang Yuan and
                  Xiwen Yao and
                  Kebing Yan and
                  Qinghua Zeng and
                  Xingxing Xie and
                  Junwei Han},
  title        = {Towards Large-Scale Small Object Detection: Survey and Benchmarks},
  journal      = {{IEEE} PAMI},
  volume       = {45},
  number       = {11},
  pages        = {13467--13488},
  year         = {2023},
}

@inproceedings{crowd_count1,
  title={Point-query quadtree for crowd counting, localization, and more},
  author={Liu, Chengxin and Lu, Hao and Cao, Zhiguo and Liu, Tongliang},
  booktitle={ICCV},
  publisher = {{IEEE}},
  pages={1676--1685},
  year={2023}
}

@inproceedings{crowd_count2,
  title={Gramformer: learning crowd counting via graph-modulated transformer},
  author={Lin, Hui and Ma, Zhiheng and Hong, Xiaopeng and Shangguan, Qinnan and Meng, Deyu},
  booktitle={AAAI},
  volume={38},
  number={4},
  pages={3395--3403},
  year={2024}
}

@article{densefuse,
  author       = {Hui Li and
                  Xiao{-}Jun Wu},
  title        = {DenseFuse: {A} Fusion Approach to Infrared and Visible Images},
  journal      = {{IEEE} TIP},
  volume       = {28},
  number       = {5},
  pages        = {2614--2623},
  year         = {2019},
}

@article{rfn-nest,
  title={RFN-Nest: An end-to-end residual fusion network for infrared and visible images},
  author={Li, Hui and Wu, Xiao-Jun and Kittler, Josef},
  journal={Inf. Fusion},
  volume={73},
  pages={72--86},
  year={2021},
  publisher={Elsevier}
}

@article{u2fusion,
  title={U2Fusion: A unified unsupervised image fusion network},
  author={Xu, Han and Ma, Jiayi and Jiang, Junjun and Guo, Xiaojie and Ling, Haibin},
  journal={IEEE PAMI},
  volume={44},
  number={1},
  pages={502--518},
  year={2020},
  publisher={IEEE}
}

@article{lrrnet,
  title={Lrrnet: A novel representation learning guided fusion network for infrared and visible images},
  author={Li, Hui and Xu, Tianyang and Wu, Xiao-Jun and Lu, Jiwen and Kittler, Josef},
  journal={IEEE PAMI},
  volume={45},
  number={9},
  pages={11040--11052},
  year={2023},
  publisher={IEEE}
}

@inproceedings{mmdrfuse,
  author       = {Yanglin Deng and
                  Tianyang Xu and
                  Chunyang Cheng and
                  Xiao{-}Jun Wu and
                  Josef Kittler},
  title        = {MMDRFuse: Distilled Mini-Model with Dynamic Refresh for Multi-Modality
                  Image Fusion},
  booktitle    = {ACM MM},
  pages        = {7326--7335},
  publisher    = {{ACM}},
  year         = {2024},
}

@article{swinfusion,
  title={SwinFusion: Cross-domain long-range learning for general image fusion via swin transformer},
  author={Ma, Jiayi and Tang, Linfeng and Fan, Fan and Huang, Jun and Mei, Xiaoguang and Ma, Yong},
  journal={IEEE/CAA JAS},
  volume={9},
  number={7},
  pages={1200--1217},
  year={2022},
  publisher={IEEE}
}

@article{AFTfusion,
  author       = {Zhihao Chang and
                  Zhixi Feng and
                  Shuyuan Yang and
                  Quanwei Gao},
  title        = {{AFT:} Adaptive Fusion Transformer for Visible and Infrared Images},
  journal      = {{IEEE} TIP},
  volume       = {32},
  pages        = {2077--2092},
  year         = {2023},
}

@article{tgfuse,
  title={TGFuse: An infrared and visible image fusion approach based on transformer and generative adversarial network},
  author={Rao, Dongyu and Xu, Tianyang and Wu, Xiao-Jun},
  journal={{IEEE} TIP},
  year={2023},
  publisher={IEEE}
}

@article{fusiongan,
  author       = {Jiayi Ma and
                  Wei Yu and
                  Pengwei Liang and
                  Chang Li and
                  Junjun Jiang},
  title        = {FusionGAN: {A} generative adversarial network for infrared and visible
                  image fusion},
  journal      = {Inf. Fusion},
  volume       = {48},
  pages        = {11--26},
  year         = {2019},

}

@inproceedings{tardal,
  title={Target-aware dual adversarial learning and a multi-scenario multi-modality benchmark to fuse infrared and visible for object detection},
  author={Liu, Jinyuan and Fan, Xin and Huang, Zhanbo and Wu, Guanyao and Liu, Risheng and Zhong, Wei and Luo, Zhongxuan},
  booktitle={CVPR},
  publisher = {{IEEE}},
  pages={5802--5811},
  year={2022}
}

@article{mufusion,
  title={MUFusion: A general unsupervised image fusion network based on memory unit},
  author={Cheng, Chunyang and Xu, Tianyang and Wu, Xiao-Jun},
  journal={Inf. Fusion},
  volume={92},
  pages={80--92},
  year={2023},
  publisher={Elsevier}
}

@inproceedings{defusion,
  title={Fusion from decomposition: A self-supervised decomposition approach for image fusion},
  author={Liang, Pengwei and Jiang, Junjun and Liu, Xianming and Ma, Jiayi},
  booktitle={ECCV},
  pages={719--735},
  year={2022},
  organization={Springer}
}

@inproceedings{cddfuse,
  title={Cddfuse: Correlation-driven dual-branch feature decomposition for multi-modality image fusion},
  author={Zhao, Zixiang and Bai, Haowen and Zhang, Jiangshe and Zhang, Yulun and Xu, Shuang and Lin, Zudi and Timofte, Radu and Van Gool, Luc},
  booktitle={CVPR},
  publisher = {{IEEE}},
  pages={5906--5916},
  year={2023}
}

@inproceedings{emma,
  author       = {Zixiang Zhao and
                  Haowen Bai and
                  Jiangshe Zhang and
                  Yulun Zhang and
                  Kai Zhang and
                  Shuang Xu and
                  Dongdong Chen and
                  Radu Timofte and
                  Luc Van Gool},
  title        = {Equivariant Multi-Modality Image Fusion},
  booktitle    = {CVPR},
  pages        = {25912--25921},
  publisher    = {{IEEE}},
  year         = {2024},
}

@inproceedings{restormer,
  author       = {Syed Waqas Zamir and
                  Aditya Arora and
                  Salman Khan and
                  Munawar Hayat and
                  Fahad Shahbaz Khan and
                  Ming{-}Hsuan Yang},
  title        = {Restormer: Efficient Transformer for High-Resolution Image Restoration},
  booktitle    = {CVPR},
  pages        = {5718--5729},
  publisher    = {{IEEE}},
  year         = {2022},
}

@inproceedings{ddfm,
  title={DDFM: denoising diffusion model for multi-modality image fusion},
  author={Zhao, Zixiang and Bai, Haowen and Zhu, Yuanzhi and Zhang, Jiangshe and Xu, Shuang and Zhang, Yulun and Zhang, Kai and Meng, Deyu and Timofte, Radu and Van Gool, Luc},
  booktitle={ICCV},
  publisher = {{IEEE}},
  pages={8082--8093},
  year={2023}
}

@inproceedings{cyclegan,
  author       = {Jun{-}Yan Zhu and
                  Taesung Park and
                  Phillip Isola and
                  Alexei A. Efros},
  title        = {Unpaired Image-to-Image Translation Using Cycle-Consistent Adversarial
                  Networks},
  booktitle    = {ICCV},
  pages        = {2242--2251},
  publisher    = {{IEEE}},
  year         = {2017},
}

@inproceedings{unpaired_1,
  author       = {Andreas Lugmayr and
                  Martin Danelljan and
                  Radu Timofte},
  title        = {Unsupervised Learning for Real-World Super-Resolution},
  booktitle    = {ICCV},
  pages        = {3408--3416},
  publisher    = {{IEEE}},
  year         = {2019},
}

@inproceedings{unpaired_2,
  author       = {Sangyun Lee and
                  Sewoong Ahn and
                  Kwangjin Yoon},
  title        = {Learning Multiple Probabilistic Degradation Generators for Unsupervised
                  Real World Image Super Resolution},
  booktitle    = {ECCV},
  volume       = {13802},
  pages        = {88--100},
  publisher    = {Springer},
  year         = {2022},
}

@inproceedings{unpaired_3,
  author       = {Yuntong Ye and
                  Yi Chang and
                  Hanyu Zhou and
                  Luxin Yan},
  title        = {Closing the Loop: Joint Rain Generation and Removal via Disentangled
                  Image Translation},
  booktitle    = {CVPR},
  pages        = {2053--2062},
  publisher    = {{IEEE}},
  year         = {2021},
}

@inproceedings{unpaired_4,
  author       = {Qian Ning and
                  Jingzhu Tang and
                  Fangfang Wu and
                  Weisheng Dong and
                  Xin Li and
                  Guangming Shi},
  title        = {Learning Degradation Uncertainty for Unsupervised Real-world Image
                  Super-resolution},
  booktitle    = {IJCAI},
  pages        = {1261--1267},
  publisher    = {ijcai.org},
  year         = {2022},
}

@article{theorm_unpaired,
  author       = {Dihan Zheng and
                  Xiaowen Zhang and
                  Kaisheng Ma and
                  Chenglong Bao},
  title        = {Learn From Unpaired Data for Image Restoration: {A} Variational Bayes
                  Approach},
  journal      = {{IEEE} PAMI},
  volume       = {45},
  number       = {5},
  pages        = {5889--5903},
  year         = {2023},
}

@article{msrs,
  author       = {Linfeng Tang and
                  Jiteng Yuan and
                  Hao Zhang and
                  Xingyu Jiang and
                  Jiayi Ma},
  title        = {PIAFusion: {A} progressive infrared and visible image fusion network
                  based on illumination aware},
  journal      = {Inf. Fusion},
  volume       = {83-84},
  pages        = {79--92},
  year         = {2022},
}

@inproceedings{roadscene,
  author       = {Han Xu and
                  Jiayi Ma and
                  Zhuliang Le and
                  Junjun Jiang and
                  Xiaojie Guo},
  title        = {FusionDN: {A} Unified Densely Connected Network for Image Fusion},
  booktitle    = {AAAI},
  pages        = {12484--12491},
  publisher    = {{AAAI}},
  year         = {2020},
}

@article{tno,
title = {The TNO Multiband Image Data Collection},
journal = {Data in Brief},
author={Toet, Alexander},
volume = {15},
pages = {249-251},
year = {2017},
}

@inproceedings{llvip,
  author       = {Xinyu Jia and
                  Chuang Zhu and
                  Minzhen Li and
                  Wenqi Tang and
                  Wenli Zhou},
  title        = {{LLVIP:} {A} Visible-infrared Paired Dataset for Low-light Vision},
  booktitle    = {ICCV Workshops},
  pages        = {3489--3497},
  publisher    = {{IEEE}},
  year         = {2021},
}

@article{metrics,
  author       = {Jiayi Ma and
                  Yong Ma and
                  Chang Li},
  title        = {Infrared and visible image fusion methods and applications: {A} survey},
  journal      = {Inf. Fusion},
  volume       = {45},
  pages        = {153--178},
  year         = {2019},

}

@article{fusionbooster,
  title={Fusionbooster: A unified image fusion boosting paradigm},
  author={Cheng, Chunyang and Xu, Tianyang and Wu, Xiao-Jun and Li, Hui and Li, Xi and Kittler, Josef},
  journal={IJCV},
  volume={133},
  number={5},
  pages={3041--3058},
  year={2025},
  publisher={Springer}
}

@inproceedings{metafusion,
  author       = {Wenda Zhao and
                  Shigeng Xie and
                  Fan Zhao and
                  You He and
                  Huchuan Lu},
  title        = {MetaFusion: Infrared and Visible Image Fusion via Meta-Feature Embedding
                  from Object Detection},
  booktitle    = {CVPR},
  pages        = {13955--13965},
  publisher    = {{IEEE}},
  year         = {2023},
}

@inproceedings{pix2pix,
  title={Image-to-image translation with conditional adversarial networks},
  author={Isola, Phillip and Zhu, Jun-Yan and Zhou, Tinghui and Efros, Alexei A},
  booktitle={CVPR},
  pages={1125--1134},
  publisher = {{IEEE}},
  year={2017}
}

@inproceedings{unpaire_restoration1,
  title={Unpaired deep image dehazing using contrastive disentanglement learning},
  author={Chen, Xiang and Fan, Zhentao and Li, Pengpeng and Dai, Longgang and Kong, Caihua and Zheng, Zhuoran and Huang, Yufeng and Li, Yufeng},
  booktitle={ECCV},
  pages={632--648},
  year={2022},
  organization={Springer}
}

@inproceedings{unpaire_restoration2,
  title={Odcr: Orthogonal decoupling contrastive regularization for unpaired image dehazing},
  author={Wang, Zhongze and Zhao, Haitao and Peng, Jingchao and Yao, Lujian and Zhao, Kaijie},
  booktitle={CVPR},
  pages={25479--25489},
  publisher = {{IEEE}},
  year={2024}
}

@inproceedings{road,
  title={Fusiondn: A unified densely connected network for image fusion},
  author={Xu, Han and Ma, Jiayi and Le, Zhuliang and Jiang, Junjun and Guo, Xiaojie},
  booktitle={AAAI},
  volume={34},
  number={07},
  pages={12484--12491},
  year={2020}
}

@article{crossfuse,
title = {CrossFuse: A novel cross attention mechanism based infrared and visible image fusion approach},
journal = { Inf. Fusion},
volume = {103},
pages = {102147},
year = {2024},
author = {Hui Li and Xiao-Jun Wu},
}

@article{cthie,
  author       = {Huafeng Li and
                  Junyu Liu and
                  Yafei Zhang and
                  Yu Liu},
  title        = {A Deep Learning Framework for Infrared and Visible Image Fusion Without
                  Strict Registration},
  journal      = {IJCV},
  volume       = {132},
  number       = {5},
  pages        = {1625--1644},
  year         = {2024},
  publisher={Springer}
}

@article{dcinn,
  title={A general paradigm with detail-preserving conditional invertible network for image fusion},
  author={Wang, Wu and Deng, Liang-Jian and Ran, Ran and Vivone, Gemine},
  journal={IJCV},
  volume={132},
  number={4},
  pages={1029--1054},
  year={2024},
  publisher={Springer}
}

@article{freefusion,
  title={FreeFusion: Infrared and Visible Image Fusion via Cross Reconstruction Learning},
  author={Zhao, Wenda and Cui, Hengshuai and Wang, Haipeng and He, You and Lu, Huchuan},
  journal={{IEEE} PAMI},
  year={2025},
  publisher={IEEE}
}

@inproceedings{sage,
  title={Every SAM Drop Counts: Embracing Semantic Priors for Multi-Modality Image Fusion and Beyond},
  author={Wu, Guanyao and Liu, Haoyu and Fu, Hongming and Peng, Yichuan and Liu, Jinyuan and Fan, Xin and Liu, Risheng},
  booktitle={CVPR},
  pages={17882--17891},
  publisher = {{IEEE}},
  year={2025}
}

@inproceedings{gifnet,
  title={One Model for ALL: Low-Level Task Interaction Is a Key to Task-Agnostic Image Fusion},
  author={Cheng, Chunyang and Xu, Tianyang and Feng, Zhenhua and Wu, Xiaojun and Tang, Zhangyong and Li, Hui and Zhang, Zeyang and Atito, Sara and Awais, Muhammad and Kittler, Josef},
  booktitle={CVPR},
  pages={28102--28112},
  publisher = {{IEEE}},
  year={2025}
}

@article{song2025,
  title={RefineFuse: an end-to-end network for multi-scale refinement fusion of multi-modality images},
  author={Song, Chengcheng and Li, Hui and Xu, Tianyang and Wu, Xiao-Jun and Kittler, Josef},
  journal={Visual Intelligence},
  volume={3},
  number={1},
  pages={16},
  year={2025},
  publisher={Springer}
}

@article{xie2024,
  title={Fusionmamba: Dynamic feature enhancement for multimodal image fusion with mamba},
  author={Xie, Xinyu and Cui, Yawen and Tan, Tao and Zheng, Xubin and Yu, Zitong},
  journal={Visual Intelligence},
  volume={2},
  number={1},
  pages={37},
  year={2024},
  publisher={Springer}
}
}
\clearpage

\end{document}